\documentclass{amsart}
\usepackage{times}
\usepackage{graphicx}
\usepackage{latexsym}
\usepackage{url}
\usepackage{multirow}

\usepackage{amsmath}
\usepackage{amsfonts} 
\usepackage{amssymb,amsmath}
\usepackage[T1]{fontenc}

\usepackage{amsthm}

{\itshape}{\rmfamily}
{\itshape}{\rmfamily}
{\bfseries}{\itshape}
\newtheorem{definition}{Definition}{\bfseries}{\itshape}
{\itshape}{\rmfamily}
{\itshape}{\rmfamily}
{\bfseries}{\itshape}
{\itshape}{\rmfamily}
{\itshape}{\rmfamily}
{\bfseries}{\itshape}
{\itshape}{\rmfamily}
{\itshape}{\rmfamily}
{\itshape}{\rmfamily}

\begin{document}

\title{Language-Constraint Reachability Learning in Probabilistic Graphs}

\author{Claudio Taranto \and Nicola Di Mauro \and Floriana Esposito}

\maketitle

\begin{abstract}
The probabilistic  graphs framework  models the uncertainty  inherent in  real-world domains by  means of 
probabilistic edges whose value  quantifies the likelihood of the edge existence  or the strength of
the link it represents.  
The goal of this paper is to provide a learning method to compute the most likely relationship between
two nodes in a  framework based on probabilistic graphs. In particular,  given a probabilistic graph
we  adopted the  language-constraint  reachability method  to  compute the  probability of  possible
interconnections that may exists between two nodes.  
Each  of these connections may  be viewed  as feature, or  a factor,  between the two  nodes and  the corresponding
probability  as its  weight.  Each  observed link  is  considered as  a  positive  instance for  its
corresponding link  label.  Given the training set  of observed links a  L2-regularized Logistic Regression  has
been adopted  to learn a model  able to predict unobserved  link labels.  The experiments  on a real
world collaborative filtering problem proved that the proposed approach achieves better results than
that obtained adopting classical methods. 
\end{abstract}

\section{INTRODUCTION}
Over the last few years the extension  of graph structures with uncertainty has become an important
research      topic~\cite{Potamias01,Zou10,Zou10a,DBLP:conf/icwsm/PfeifferN11},      leading      to
\emph{probabilistic  graph}\footnote{The  names   \emph{probabilistic  graphs}  and  \emph{uncertain
    graphs}  are usually  used  to refer  the  same framework.}  model.  Probabilistic graphs  model
uncertainty  by means  of probabilistic  edges whose  value quantifies  the likelihood  of  the edge
existence or the strength of the link it represents.  
One of the main issues in probabilistic graphs is how to compute the connectivity of the network.  
The network reliability problem~\cite{Colbourn87} is a generalization of the pairwise reachability, 
in which the goal is to determine the probability that all pairs of nodes are reachable from one another. 
Unlike a deterministic graph in which the reachability function is a binary value function indicating whether 
or not there is a path connecting  two nodes,
in the case of probabilistic graphs the function assumes probabilistic values.

The concept of \emph{reachability} in probabilistic graphs is used, along with its specialization, as a
tool to compute how two nodes in the graph are likely to be connected. 
Reachability plays an important role in wide range of applications, such as in 
peer-to-peer networks~\cite{Clarke:2001:FDA:371931.371977,Pandurangan:2001:BLP:874063.875584}, 
 for 
probabilistic-routing problem~\cite{Biswas:2005:EOM:1080091.1080108,citeulike:2688183}, 
in road network~\cite{Hua:2010:PPQ:1739041.1739084}, 
 and in trust analysis in social networks~\cite{Swamynathan:2008:SNI:1397735.1397737}.
As adopted in these  works, reachability is quite similar to the  general concept of \emph{link prediction}~\cite{DBLP:journals/sigkdd/GetoorD05a},
whose task may be formalized as follows. Given a networked structure $(V,E)$ made up of a set of
data instances $V$  and set of observed links $E$  among some nodes in $V$,  the task corresponds to
predict how likely should exist an unobserved link between two nodes in the network. 

The  extension to  probabilistic  graphs adds  an  important ingredient  that  should be  adequately
exploited. The  key difference with respect  to classical link  prediction methods is that  here the
observed  connections  between  two nodes  cannot  be  considered  always  true, and  hence  methods
exploiting probabilistic links are needed. 
Link prediction can be specialized into link  existence prediction, where one wants to asses whether
two nodes  should be connected, and  link classification, where  one is interested in  computing the
most likely relationship existing between two nodes. 

The goal of this paper is to provide a learning
method to compute the most likely relationship between two nodes in probabilistic graphs. 
In  particular,  given a  probabilistic  graph  we adopted  the  reachability  tool  to compute  the
probability  of some possible  interconnections that  may exists  between two  nodes. Each  of these
connections may  be viewed  as a feature, or  a factor,  between the two  nodes and  the corresponding
probability  as its  weight.  Each  observed labeled link  is  considered as  a  positive  instance  for  its
corresponding link  label. In  particular, the  link label corresponds  to the  value of  the output
variable  $y_i$, and  the  features  between the  two  nodes, computed with the reachability tool,  correspond  to  the  components of  the
corresponding  vector  $\mathbf  x_i$.  Given  the  training  set  $\mathcal  D  =  \{(\mathbf  x_i,
y_i)\}_{i=1}^n$, obtained from $n$ observed links,   a L2-regularized Logistic Regression  has  been  adopted to  learn a  model  to be  used to  predict
unobserved link labels.

The application domain we chosen corresponds to the problem of recommender systems~\cite{reference/rsh/DesrosiersK11}, where the aim is
to predict the unknown rating between an user  and an item.  The experiments on a real-world dataset 
prove that the proposed  approach achieves better results than that obtained  with models induced by
Singular Value Decomposition (SVD)~\cite{Dartmouth:TR98-338} on the user-item ratings matrix, representing one of the
best recent methods for this kind of task~\cite{Koren:2008:FMN:1401890.1401944}.
The paper is organized as follows: Section~\ref{uncgra} presents the probabilistic graphs framework, Section~\ref{learning} describes the proposed link classification approach, Section~\ref{relworks} describes related works, 
and finally Section~\ref{exp} shows the experimental results.


\section{PROBABILISTIC GRAPHS}
\label{uncgra}
Let $G = (V, E)$,  be a graph where $V$ is a collection of nodes and $E \in V \times V$ is the set
of edges, or relationships, between the nodes. 

\begin{definition}[Probabilistic graph]
A \emph{probabilistic  graph} is a system  $G = (V,E,$ $\Sigma,  l_V, l_E, P_e)$, where  $(V,E)$ is an
undirected graph, $V$ is the set of nodes, $E$ is the set of edges, $\Sigma$ is a set of labels,
$l_V : V \rightarrow \Sigma$ is a function assigning labels to nodes,
$l_E : E \rightarrow \Sigma$ is a function assigning labels to the edges, and 
$P_e : E \rightarrow [0,1]$ is a function assigning \emph{existence probability} values to the
edges.
\end{definition}
The existence  probability $P_e(a)$ of an edge  $a = (u,v) \in  E$ is the probability  that the edge
$a$, between $u$ and $v$, can exist in the graph. 
A  particular case  of probabilistic  graph is  the \emph{discrete  graph}\footnote{Sometimes called
  \emph{certain graph}.},  where binary edges between nodes  represent the presence or  absence of a
relationship between them, i.e., the existence probability value on all observed edges is 1. 

The \emph{possible  world semantics}  is usually used for probabilistic  graphs. We can  imagine a
probabilistic graph $G$ as a sampler of worlds, where each world is an instance of $G$. A discrete 
graph $G'$  is sampled  from $G$  according to the  probability distribution  $P_e$, denoted  as $G'
\sqsubseteq  G$, when  each edge  $a \in  E$ is  selected to  be an  edge of  $G'$  with probability
$P_e(a)$. 
Edges labeled with probabilities are treated as mutually independent random variables indicating
whether or not the corresponding edge belongs to a discrete graph.

Assuming independence among edges, the probability distribution over
discrete graphs $G' =(V,E') \sqsubseteq G = (V,E)$  is given by
\begin{equation}
 P(G'|G) = \prod_{a \in E'} P_e(a) \prod_{a \in E\setminus E'} (1- P_e(a)).     
 \end{equation}

\begin{definition}[Simple path]
Given a probabilistic  graph $G$, a \emph{simple path} of a  length $k$ from $u$ to $v$  in $G$ is a
sequence  of edges  $p_{u,v} =  \langle e_1,  e_2, \ldots  e_k \rangle$,  such that  $e_1=(u, v_1)$,
$e_k=(v_{k_1},v)$, and 
$e_i=(v_{i-1},v_i)$ for $1 < i < k$, and all nodes in the path are distinct.
\end{definition}

Given $G$ a probabilistic graph, and $p_{s,t} =  \langle e_1, e_2, \ldots e_k \rangle$ a simple path
in $G$ from node $s$ to node $t$, $l(p_{s,t}) = 
l_E(e_1)l_E(e_2)  \cdots  l_E(e_k)$  denotes  the  concatenation  of the  labels  of  all  edges  in
$p_{s,t}$.  In order to give the following definition, we recall that given a \emph{context free grammar} (CFG) $\mathcal C$ a string of
terminals $s$ is derivable from $\mathcal C$ iff $s \in L(\mathcal C)$, where $L(\mathcal C)$ is the
language generated from $\mathcal C$.

\begin{definition}[Language constrained simple path]
Given a probabilistic graph $G$ and a context free grammar $\mathcal C$, a \emph{language constrained
simple path} is a simple path $p$ such that $l(p) \in L(\mathcal C)$.
\end{definition}

\subsection{Inference}

Given a probabilistic graph $G$ a main task corresponds to compute the probability that there exists
a simple path between  two nodes $u$ and $v$, that is, querying for  the probability that a randomly
sampled  discrete  graph   contains  a  simple  path  between  $u$  and   $v$.  More  formally,  the
\emph{existence  probability}  $P_e(q|G)$  of  a  simple  path $q$  in  a  probabilistic  graph  $G$
corresponds to the marginal $P(G'|G)$ with respect to $q$:
\begin{equation}
  P_e(q|G) = \sum_{G' \sqsubseteq G} P(q|G') \cdot P(G'|G)
  \label{exactinf}
\end{equation} 
where $P(q|G')=1$ if there exits the simple path $q$ in $G'$, and $P(q|G')=0$ otherwise.
In other words, the existence probability of the  simple path $q$ is the probability that the simple
path $q$ exists in a randomly sampled discrete graph.

\begin{definition}[Language constrained simple path probability]
Given a probabilistic graph $G$ and a context free grammar $\mathcal C$, the \emph{language constrained
simple path probability} of $L(\mathcal C)$ is 
\begin{equation}
    P(L(\mathcal C)|G) = \sum_{G' \sqsubseteq G} P(q|G',L(\mathcal C)) \cdot P(G'|G)
    \label{languagepath}
\end{equation} 
where $P(q|G',L(\mathcal C)=1$ if there exists a simple path $q$ in $G'$ such that $l(q) \in L(\mathcal C)$, and $P(q|G',L(\mathcal C))=0$ otherwise.
\end{definition}
In particular, the previous definition give us the possibility to compute the probability of a set of simple path queries
fulfilling  the structure  imposed by  a context  free grammar.  In this  way we  are  interested in
discrete graphs that contain at least one simple path belonging to the language corresponding to the
given grammar.

Computing  the existence  probability  directly using  (\ref{exactinf})  or (\ref{languagepath})  is
intensive and intractable for 
large graphs since the number of discrete graphs to be checked is exponential in the number of
probabilistic edges. It involves computing the existence of the simple path in every discrete graph and 
accumulating their probability.
A natural way to overcome the intractability of computing the existence probability of a simple path is to
approximate  it using  a  Monte Carlo  sampling  approach~\cite{Jin11}: 1)  we  sample $n$  possible
discrete graphs, $G_1, G_2, 
\ldots G_n$ from $G$ by sampling edges uniformly at random according to
their edge probabilities; and 2) we check if the simple path exists in each
sampled graph $G_i$. This process provides the following basic sampling estimator for $ P_e(q|G)$:
\begin{equation}
\label{ref:app} 
 \widehat{P_e}(q|G) = \frac{\sum_{i=1}^n P(q|G_i) }{n}
\end{equation}
  
Note that is not necessary to sample all edges to check whether the graph contains the path. For
instance, assuming to use an iterative depth first search procedure to check the path existence.
When a node is just visited, we will sample all its adjacent edges and pushing them into the stack
used by the iterative procedure. We will stop the procedure either when the target node is reached or
when the stack is empty (non existence).

\section{LINK CLASSIFICATION}
\label{learning}

After having defined the probabilistic graph, now we can adopt language constrained simple paths 
in order to extract probabilistic features to describe the link between two nodes in the graph. 

Given  a probabilistic  graph $G$,  with the  set $V$  of nodes  and the  set $E$  of edges,  and $Y
\subseteq \Sigma$ a  set of edge labels, we have a  set of edges $D \subseteq E$ such that
for each element $e \in D$: $l_E(e) \in Y$. In particular $D$ represents the set of observed links whose 
label belongs to the set $Y$. Given
the set of training links  $D$ and the set of labels $Y$ we want to  learn a model able to correctly
classify unobserved links. 

\subsection{Query based classification}
\label{sec:qc}
A way to solve the classification task can be that of using a language based classification approach.
Given an unobserved edge $e_i = (u_i,v_i)$,  in order to predict its class $\widehat{y_i}
\in Y$ we can solve the following maximization problem: 
\begin{equation}
    \widehat{y_i} = \arg \max_j P(q_j(u_i,v_i)|G),
    \label{predicting}
\end{equation}
where $q_j(u_i,v_i)$ is the unknown link with label $q_j \in Y$ between the nodes $u_i$ and $v_i$. 
In particular,
the maximization problem  corresponds to compute the link  prediction for each $q_j \in  Y$ and then
choosing that label with maximum likelihood.
The previous link prediction task is based on querying the probability of some language constrained simple path.
In particular, predicting the probability of the label $q_j$ as $P(q_j(u_i,v_i)|G)$ in (\ref{predicting}) 
corresponds  to compute  the  probability $P(q|G)$  for  a query  path in  a  language $L_j$,  i.e.,
computing $P(L_j|G)$ as in (\ref{languagepath}):
\begin{equation}
    \label{predicting1}
    \widehat{y_j} = \arg \max_j P(q_j(u_i,v_i)|G) \approx \arg \max_j P(L_j|G).
\end{equation}

\subsection{Feature based classification}

The previous query based classification approach  consider the languages  used to compute  the (\ref{predicting1}) as independent  form each
other without  considering any correlation between them.  A more interesting approach  that we want
investigate in this paper is to learn  from the probabilistic graph a linear model of classification
combining the prediction of each language constrained simple path.

In particular,  given an  edge $e$  and a set  of $k$  languages $\mathcal L = \{L_1,\ldots,L_k\}$,  we can  generate $k$
real valued features $x_i$  where $x_i = P(L_i|G)$, $1 \leq i \leq k$.  The original training set of
observed  links  $D$  can  hence be  transformed  into  the  set  of  instances  $\mathcal  D  =  \{(\mathbf
x_i,y_i)\}_{i=1,\ldots,n}$,  where $\mathbf  x_i$ is  a  $k$-component vector  of
features $x_{ij} \in [0,1]$, and $y_i$ is the class label of the corresponding example $\mathbf x_i$.  

\subsubsection{L2-regularized Logistic Regression}
\label{sec:learn}
Linear classification  represents one  of the  most promising
learning technique for  problems with a huge number  of instances and features aiming  at learning a
weight vector $\mathbf w$ as a model. L2-regularized Logistic Regression belongs to the class of linear
classifier and solves the following unconstrained optimization problem:
\begin{equation}
\min_{\mathbf w} f(\mathbf w) = \left ( \frac{\mathbf{w}^T\mathbf{w}}{2} + C \sum_{i=1}^n \log (1+\exp(-y_i \mathbf w^T \mathbf x_i)) \right ),
\end{equation}
where $\log  (1+\exp(-y_i \mathbf w^T \mathbf  x_i))=\xi(\mathbf  w; \mathbf x_i,y_i)$ denotes the
specific loss function, $\frac{1}{2} \mathbf{w}^T\mathbf{w}$ is the regularized term, and $C>0$ is a penalty parameter. The decision function corresponds to $\text{sgn}(\mathbf
w^t \mathbf x_i)$. In case of binary classification $y_i \in \{-1,+1\}$,
while for multi class problems the one vs the rest strategy can be used.

Among many  methods for training  logistic regression models,  such as iterative  scaling, nonlinear
conjugate gradient, quasi Newton, a new efficient and robust truncated Newton,  called trust
region Newton method, has been
proposed~\cite{DBLP:journals/jmlr/LinWK08}. 

In order to find the parameters $\mathbf w$ minimizing $f(\mathbf w)$ it is necessary to set the
derivative of $f(\mathbf w)$ to zero.  Denoting with $\sigma(y_i \mathbf w^T \mathbf x_i) =
(1+exp(-y_i\mathbf w^T \mathbf x_i))^{-1}$, we have:
$$\frac{\partial  f(\mathbf  w)}{\partial  \mathbf  w}  =   \mathbf  w  +  C  \sum_{i=1}^n  \left  (
  \sigma(y_i \mathbf w^T \mathbf x_i)-1 \right )y_i \mathbf x_i = 0.$$
To solve the previous score equation, the Newton method requires the Hessian matrix:
$$\frac{\partial^2 f(\mathbf  w)}{\partial \mathbf w \partial \mathbf  w^T} = \mathbf I  + C \mathbf
X^T \mathbf D \mathbf X,$$ where $\mathbf X$  is the matrix of the $\mathbf x_i$ values, $\mathbf D$
is  a diagonal  matrix  of  weights with  $i$th  diagonal element  $\sigma(y_i  \mathbf w^T  \mathbf
x_i)(1-\sigma(y_i \mathbf w^T \mathbf x_i))$, and $\mathbf I$ is the identity matrix.

The  Newton   step  is  $$\mathbf   w^{\text{new}}  \leftarrow  \mathbf  w^{\text{old}}   +  \mathbf
s^{\text{old}},$$ where $\mathbf s^{\text{old}}$ is the solution of the following linear system:
$$\frac{\partial^2  f(\mathbf  w^{\text{old})}}{\partial  \mathbf  w \partial  \mathbf  w^T}  \mathbf
  s^{\text{old}} = - \frac{\partial f(\mathbf w^{\text{old}})}{\partial \mathbf w }.$$

Instead of using this update  rule, \cite{DBLP:journals/jmlr/LinWK08} propose a robust and efficient
trust  region Newton method,  using new  rules for  updating the  trust region,  whose corresponding
algorithm has been implemented in the \textsf{LIBLINEAR}\footnote{\url{http://www.csie.ntu.edu.tw/~cjlin/liblinear
}.} system. 

\section{EXPERIMENTAL EVALUATION}
\label{exp}
The  application  domain  we chosen  to  validate  the  proposed  approach  is that  of  recommender
systems. In  some domains  both data  and probabilistic relationships  between them  are observable,
while in  other domain, like  in this used in  this paper, it  is necessary to elicit  the uncertain
relationships among the given evidence.
\subsection{Probabilistic graph creation}
\label{sec:building}

A  common approach  to elicit  probabilistic hidden  relationships between  data is  based  on using
similarity measures. 
To model the  data with a  graph we can  adopt different similarity measures  for each type  of node
involved in the  relationships. For instance we can define a  similarity measure between homogeneous
nodes and one for heterogeneous nodes.

In a recommender system we have two types of entities: the users and the items, and the only 
observed relationship  corresponds to the ratings  that a user has  assigned to a set  of items. The
goal is to predict the rating a user could assign to an object that he never rated in the past. 
In the collaborative  filtering approach there are two methods to  predict unknown rating exploiting
users or items similarity. User-oriented methods  estimate unknown ratings based on previous ratings
of similar  users, while in  item-oriented approaches ratings  are estimated using  previous ratings
given by the same user on similar items. 

Let $U$ be a set of $n$ users and $I$ a set of $m$ items. A rating $r_{ui}$ indicates the preference
degree the  user $u$ expressed  for the item  $i$, where high  values mean stronger  preference. Let
$S_u$ be the set  of items rated from user $u$. A user-based  approach predicts an unobserved rating
$\widehat{r_{ui}}$ as follows:
\begin{equation}
  \widehat{r_{ui}} = \overline{r_u} + \frac{\sum_{v \in U | i \in S_u} \sigma_u(u,v) \cdot (r_{vi} -
  \overline{r_v})}{\sum_{v \in U | i \in S_u} |\sigma_u(u,v)|} 
\end{equation}
where $\overline{r_u}$ represents the mean rating of user $u$, and $\sigma_u(u,v)$ stands for the
similarity between users $u$ and $v$, computed, for instance, using the Pearson correlation:
\begin{equation*}
    \label{pearson}
  \sigma_u(u,v) = \frac{\sum_{a \in S_u \cap S_v} (r_{ua} - \overline{r_u}) \cdot (r_{va} -
  \overline{r_v})}{\sqrt{\sum_{a \in S_u \cap S_v} (r_{ua} - \overline{r_u})^2 \sum_{a \in S_u \cap
  S_v}(r_{va} - \overline{r_v})^2}}
\end{equation*}

On the other side, item-based approaches predict the rating of a given item using the following formula:
\begin{equation}
  \widehat{r_{ui}} = \frac{\sum_{ j \in S_u | j \neq i} \sigma_i(i,j) \cdot r_{uj}}{\sum_{j \in S_u | j \neq i}
  |\sigma_i(i,j)|},
\end{equation}
where $\sigma_i(i,j)$ is the similarity between the item $i$ and $j$.

These neighbourhood approaches see each user connected to other users or consider each item related
to other items as in a network structure. In particular they rely on the direct connections among the 
entities involved in the domain. However, as recently proved, techniques able to consider complex relationships among the
entities, leveraging the information already present  in the network, involves an improvement in the processes of querying
and mining~\cite{Witsenburg11,Taranto11,taranto12iir}.


Given the set of observed ratings $\mathcal K = \{(u,i,r_{ui}) | r_{ui}$ is known$\}$, we add a node with label
\texttt{user} for each user in $\mathcal K$, and a node with label \texttt{item} for each item in
$\mathcal K$. The next step is to add the edges among the nodes. Each edge is characterized by a
label and a probability value, which should indicate the degree of similarity between the two
nodes. Two kind of connections between nodes are added. For each user $u$, we added an edge, labeled
as \texttt{simU}, between $u$ and the $k$ most similar users to $u$. The similarity between two users
$u$ and $v$ is computed adopting a weighted Pearson correlation between the items rated by both $u$ and
$v$. 
In particular, the probability of the edge \texttt{simU} connecting two users $u$ and $v$ is computed as:
\begin{eqnarray*}
    P(\texttt{simU}(u, v))= \sigma_u(u,v) \cdot w_u(u, v),
\end{eqnarray*}
where $\sigma_u(u,v)$ is the Pearson correlation between the vectors of ratings corresponding to the
set of items rated by both user $u$ and user $v$, and 
 $ w_u(u, v) = \frac {|S_u \cap S_v |}{|S_u \cup S_v| }$.

For each item $i$, we added an edge, with label \texttt{simI}, between $i$ and the most $k$
similar items to $i$.
In particular, the probability of the edge \texttt{simI} connecting the item $i$ to the item $j$ has been computed as:
\begin{eqnarray*}
    P(\texttt{simI}(i,j)) = \sigma_i(i,j) \cdot w_i(i,j),
\end{eqnarray*}
where $\sigma_i(i,j)$ is the Pearson correlation between the vectors corresponding to the histogram 
of the set of ratings for the item $i$ and the item $j$, and
   $ w_i(i, j)=\frac {| \overline{S}_i \cap \overline{S}_j|}{|\overline{S}_i \cup \overline{S}_j|}$,
where $\overline{S}_i$ is the set of users rating the item $i$.

Finally, edges with probability equal to 1, and with label $\texttt{r}_k$ between the user $u$
and the item $i$, denoting the user $u$ has rated the item $i$ with a score equal to $k$, 
are added for each element $r_{ui}$ belonging to $\mathcal K$.

\subsection{Feature construction}
\label{sec:costr}
Let  us assume that  the values  of $r_{ui}$  are discrete  and belonging  to a  set $R$.  Given the
recommender  probabilistic  graph $G$,  the  query based  classification  approach,  as reported  in
Section~\ref{sec:qc}, try to solve the  problem
$\widehat{r_{ui}} = \arg \max_j P(\texttt{r}_j(u,i)|G)$,
where $\texttt{r}_j(u,i)$ is the unknown link with label $\texttt{r}_j$ between the user $u$ and the item $i$. 
This link prediction task is based on querying the probability of some language constrained simple path.
For  instance,  a user-based  collaborative  filtering  approach may  be  obtained  by querying  the
probability of  the paths, starting from a  user node and ending  to an item node,  belonging to the
context free language (CFL) $L_i=\{\texttt{simU}^1 \texttt{r}_i^1\}$.  
In particular, predicting the probability of the rating $j$ as $P(\texttt{r}_j(u,i))$ 
corresponds to compute the probability $P(q|G)$ for a query path in $L_j$, i.e., 
$\widehat{r_{ui}} = \arg \max_j P(\texttt{r}_j(u,i)|G) \approx \arg \max_j P(L_j|G)$.

In the same way, item-based approach could
be  obtained by computing the probability of the paths belonging to the CFL $L_i=\{\texttt{r}_i^1 \texttt{simI}^1\}$.  
The power of the proposed framework gives us the possibility to construct more complex queries such
as that belonging to the CFL $L_i=\{\texttt{r}_i \texttt{simI}^n : 1 \leq n \leq 2\}$, that gives us the possibility
to explore the graph by considering not only direct connections.
Hybrid queries, such  as those belonging to  the CFL $L_i=\{\texttt{r}_i \texttt{simI}^n :  1 \leq n
\leq 2\}  \cup \{\texttt{simU}^m \texttt{r}_i^1  : 1 \leq  m \leq 2\}$,  give us the  possibility to
combine the user information with item information.

In order to use the feature based classification approach proposed in this paper we can define a set
of CFLs  $\mathcal{L}$ and then  computing for  each language $L_i  \in \mathcal L$  the probability
$P(L_i|G)$ between a given user and all the items the user rated. In particular, the set of observed
ratings $\mathcal K = \{(u,i,r_{ui}) | r_{ui}$  is known$\}$ is mapped to the training set $\mathcal
D = \{(\mathbf x_i,y_i)\}_{i=1,\ldots,n}$, where  $x_{ij}$ is the probability $P(L_j|G)$ between the
nodes $u$ and $i$, and $y_i$ is equal to $r_{ui}$.

The   proposed   link  classification   method   has   been   implemented  in   the   \texttt{Eagle}
system\footnote{\url{http://www.di.uniba.it/~claudiotaranto/eagle.html}}  that  provides  a  set  of
tools to deal with probabilistic graphs.  

\subsection{Dataset}

In     order    to     validate    the     proposed     approach    we     used    the     MovieLens
dataset\footnote{\url{http://ir.ii.uam.es/hetrec2011/datasets.html}},   made    available   by   the
GroupLens research group at University of Minnesota for the 2nd 
International Workshop on Information Heterogeneity and Fusion in Recommender Systems.
We used the MovieLens 100K version consisting of  100000 ratings (ranging from 1 to 5) regarding 943
users and 1682 movies, whose class distribution is reported in Table~\ref{data}. Each user has rated at least 20 movies and there are simple demographic info
for the users (such as age, gender, occupation, and zip code). 
The data was collected  through the MovieLens web site during the  seven-month period from September
19th, 1997 through April 22nd, 1998. In this  paper we used the ratings only without considering the
demographic information.  MovieLens 100K dataset  is divided  in 5 fold,  where each fold  present a
training data (consisting of 80000 ratings) and a test data (with 20000 ratings).

\begin{table}
\centering
\caption{MovieLens dataset class distribution.}
\label{data}
\begin{tabular}{|ccccc|}
\hline
r1 & r2 & r3 & r4 & r5\\
\hline \hline
6110 & 11370 &27145& 34174& 21201 \\
\hline
\end{tabular}
\end{table}

For each training/testing fold the validation procedure followed the following steps:
\begin{enumerate}
    \item  creating  the  probabilistic  graph  from  the training  ratings  data  set  as  reported
      Section~\ref{sec:building}; 
    \item defining a set  $\mathcal L$ of context free languages corresponding  to be used to construct  a specific set of
      features as described in Section~\ref{sec:costr}; 
    \item learning the L2-regularized Logistic Regression model; and 
    \item testing the ratings  reported in the testing data set $\mathcal  T$ by computing, for each
      pair $(u,i)  \in \mathcal T$  the predicted rating  adopting the learned classification model and  comparing the
      result with the true prediction reported in $\mathcal T$. 
\end{enumerate}

For the graph construction, edges are added using the procedure presented in Section~\ref{sec:building}, 
where we set the parameter $n=30$, indicating that  an user or a film is connected, respectively, to
30 most similar users, resp. films. 
The value of  each feature have been obtained  with the Monte Carlo inference  procedure by sampling
100 discrete graphs.

In order to  construct the set of features, we  proposed to query the paths belonging  to the set of
languages $\mathcal  L$ reported in Table~\ref{paths}.  The first language  constrained simple paths
$L_1$ corresponds  to adopt  a user-based  approach, while the  second language  $L_2$ gives  us the
possibility to apply an  item-based approach.  Then, we propose to extend  the basic languages $L_1$
and $L_2$ in order  to construct features that consider a neighbourhood  with many nested levels. In
particular, instead  of considering the direct  neighbours only, we inspect  the probabilistic graph
following a path with a maximum length of two ($L_3$ and $L_4$) and three edges ($L_6$ and $L_7$). 
Finally, we constructed hybrid features by  combining both the user-based and item-based methods and
the large neighbourhood explored with paths whose length is greater than one ($L_5$, $L_8$ and $L_9$).
We defined  two sets  of features $\mathcal{F}_1  = \{L_1,  L_2, L_3, L_4,  L_5\}$, based  on simple
languages,  and $\mathcal{F}_2  = \{  L_3, L_4,  L_5, L_6,  L_7, L_8,  L_9\}$, exploiting  more 
complex  queries. In order to learn the classification model 
as reported in Section~\ref{sec:learn}, we used the L2-regularized Logistic Regression implementation 
included in the \textsf{LIBLINEAR} system~\cite{DBLP:journals/jmlr/LinWK08}.

\begin{table}
    \caption{Language constrained simple paths used for the MovieLens dataset.}
    \centering
    \begin{tabular}{|cl|}
        \hline
        $L_1 =$ & $\{ \texttt{simU}^1 \texttt{r}_k^1 \}$\\
	$L_2 =$ & $\{ \texttt{r}_k^1 \texttt{simF}^1 \}$\\
	$L_3 =$ & $\{ \texttt{r}_k^1 \texttt{simF}^n : 1 \leq n \leq 2 \}$\\
        $L_4 =$ & $\{ \texttt{simU}^n \texttt{r}_k^1 : 1 \leq n \leq 2 \}$\\
        $L_5 =$ & $\{ \texttt{simU}^n \texttt{r}_k^1 : 1 \leq n \leq 2 \} \cup \{ \texttt{r}_k^1 \texttt{simF}^n : 1 \leq n \leq 2 \}$\\
	$L_6 =$ & $\{ \texttt{r}_k^1 \texttt{simF}^n : 1 \leq n \leq 3 \}$\\
        $L_7 =$ & $\{ \texttt{simU}^n \texttt{r}_k^1 : 1 \leq n \leq 3 \}$\\
        $L_8 =$ & $\{ \texttt{simU}^n \texttt{r}_k^1 : 1 \leq n \leq 3 \} \cup \{ \texttt{r}_k^1 \texttt{simF}^n : 1 \leq n \leq 3 \}$\\
        $L_9 =$ & $\{ \texttt{simU}^n \texttt{r}_k^1 : 1 \leq n \leq 4 \} \cup \{ \texttt{r}_k^1 \texttt{simF}^n : 1 \leq n \leq 4 \}$\\
        \hline
    \end{tabular}
    \label{paths}
\end{table}

Given a  set $\mathcal  T$ of testing  instances, the  accuracy of the  proposed framework  has been
evaluated     according     to     the     \emph{macroaveraging    mean     absolute     error     }
($MAE^M$)~\cite{Baccianella:2009:EMO:1681507.1682004}: 
\begin{equation*}
    MAE^M(\widehat{r_{ui}},\mathcal{T})= \frac{1}{k} \sum_{j=1}^k \frac{1}{|T_j|} \sum_{x_i \in T_j} |\widehat{r_{ui}} - r_{ui}|
\end{equation*}
where $T_j \subset \mathcal T$ denotes the set of test rating whose true class is $j$.

\subsection{Results}
 Table~\ref{tab:mae}  shows the  results  obtained adopting  the proposed  approach
implemented in the \texttt{Eagle} system when compared to those obtained with the RecSys SVD approach based
implementation\footnote{\url{https://github.com/ocelma/python-recsys}}.  The first  row  reports the
mean value  of the MAE$^M$  averaged on the five  folds obtained with  an SVD approach and  with the
proposed classification method as implemented in the  \texttt{Eagle} system. As we can see the error achieved
by our method is lower than that obtained by the SVD method. The results improve when we use the set
$\mathcal  F_2$ of  features.  The difference  of  the results  obtained with  the  two methods  is
statistically significant,  with a  p-value for  the t-test equal  to 0.0000023  when using  the set
$\mathcal F_1$ of features,  and equal to $0.000000509$ for the other set  of features. The last two
columns report  the results  of two  baseline methods. The  second last  column reports  the results
obtained with a system that predicts a rating adopting a uniform distribution, while the last column
reports the results of a system that uses  a categorical distribution that predicts the value $k$ of
a rating  with probability $p_k =  |D_k|/N$, where $D_k$ is  the number of ratings  belonging to the
dataset having value $k$, and $N$ is the total number of ratings.

\begin{table}
\caption{MAE$^M$ values obtained with \texttt{Eagle} and SVD on MovieLens dataset.}
\centering
\begin{tabular}{|c|ccccc|}
\hline
Fold & \texttt{SVD} & \texttt{Eagle}@$\mathcal{F}_1$ & \texttt{Eagle}@$\mathcal{F}_2$ & \texttt{U} & \texttt{C}\\
\hline \hline 
1 & 0.9021 & 0.8424 & 0.8255 & & \\
2 & 0.9034 & 0.8332 & 0.8279& & \\
3 & 0.9111 & 0.8464 & 0.8362& & \\
4 & 0.9081 & 0.8527 & 0.8372& & \\
5 & 0.9159 & 0.8596 & 0.8502&  & \\
\hline
Mean & 0.908$\pm$0.006 & 0.847$\pm$0.01 & 0.835$\pm$0.01 & 1.6 & 1.51\\
\hline
p-value & & 2.3E-6 & 5.09E-7 & & \\
\hline
\end{tabular}\\
\hfill
\label{tab:mae}
\end{table}

In Table~\ref{tab:mae1}  we can see the  errors committed by each  method on each  rating class. The
rows for the  methods \texttt{U} and \texttt{C} report  the mean of the MAE$^M$ value  for each fold
using a system adopting a uniform or a categorical distribution. The dataset is not balanced as
reported in the Table~\ref{data}. As we can see both the \texttt{SVD} and the \texttt{Eagle} system adhere more to the
categorical distribution proving that they are able to recognize the unbalanced distribution of the dataset

\begin{table}
\caption{MAE$^M$ values for each class obtained with \texttt{Eagle} and SVD on MovieLens dataset.}
\centering
\begin{tabular}{|c|c|c c c c c|}
\hline
Fold & Method & r1 & r2 & r3 & r4 & r5\\
\hline
 & \texttt{SVD} & 1.58 & 1.04 & 0.56 & 0.44 & 0.86\\
1& \texttt{Eagle}@$\mathcal F_1$ & 1.11 & 0.76 & 0.69 & 0.61 & 1.02\\
 & \texttt{Eagle}@$\mathcal F_2$ & 1.03 & 0.75 & 0.71 & 0.63 & 0.99\\
\hline
 & \texttt{SVD} & 1.60 & 1.04 & 0.55 & 0.43 & 0.87\\
2 & \texttt{Eagle}@$\mathcal F_1$ & 1.11 & 0.77 & 0.67 & 0.58 & 1.02\\
 & \texttt{Eagle}@$\mathcal F_2$ & 1.05 & 0.77 & 0.68 & 0.60 & 1.00\\
\hline
 & \texttt{SVD}  & $1.65$& $0.99$& $0.55$ & $0.45$& $0.89$\\
3 & \texttt{Eagle}@$\mathcal F_1$   & $1.20$& $0.74$& $0.66$ & $0.60$& $1.02$\\
 & \texttt{Eagle}@$\mathcal F_2$ & $1.15$& $0.74$& $0.66$ & $0.64$& $0.98$\\
\hline
 & \texttt{SVD}  & $1.62$& $1.04$& $0.53$ & $0.45$& $0.87$\\
4 & \texttt{Eagle}@$\mathcal F_1$   & $1.21$& $0.75$& $0.66$ & $0.59$& $1.02$\\
 & \texttt{Eagle}@$\mathcal F_2$ & $1.14$& $0.75$& $0.66$ & $0.60$& $1.02$\\
\hline
 & \texttt{SVD}  & $1.65$& $1.03$& $0.55$ & $0.44$& $0.89$\\
5 & \texttt{Eagle}@$\mathcal F_1$   & $1.19$& $0.76$& $0.66$ & $0.63$& $1.03$\\
 & \texttt{Eagle}@$\mathcal F_2$ & $1.16$& $0.75$& $0.67$ & $0.64$& $1.01$\\
\hline
\hline 
\multirow{5}{*}{Mean} & \texttt{U} & 2.0 & 1.4 & 1.2 & 1.4 & 2.0\\
& \texttt{C} & 2.53 & 1.65 & 1.00 & 0.89 & 1.47\\
& \texttt{SVD} & 1.62 & 1.03 & 0.55 & 0.44 & 0.88\\
& \texttt{Eagle}@$\mathcal F_1$ & 1.16& 0.76& 0.67 & 0.60 & 1.02\\
& \texttt{Eagle}@$\mathcal F_2$ & 1.11& 0.75& 0.68 & 0.62 & 1.00\\
\hline 
\end{tabular}
\label{tab:mae1}
\end{table}

\section{RELATED WORKS}
\label{relworks}
In~\cite{Potamias01} the authors provide a list of alternative shortest-path distance measures for probabilistic graphs in order to discover 
the $k$ closest nodes to a given node. Their work is related to the that of stochastic shortest path problem  that deals with the computing
of the probability density function of the shortest path length for a pair of nodes~\cite{Frank_1969}.
They provide a scalable solution for the k-NN problem by using a direct sampling approach that  approximates the shortest-path 
probability between two nodes adopting a sampling of $n$ possible discrete
graphs from the probabilistic graph and hence computing the shortest path distance in each sampled discrete graph.
In~\cite{Dantzig98}, the problem of
finding a shortest path on a probabilistic graph is addressed by transforming each edge probability to its expected value and then
running the Dijkstra algorithm. 

Authors in~\cite{Jin11} investigated a more generalized and informative  distance-constraint reachability (DCR) query problem: given 
two nodes $s$ and $t$ in an probabilistic  graph $G$, the aim is to compute the probability that the distance form $s$ to $t$ is less than 
or equal to $d$. They show that the simple reachability problem without constraint becomes a special
case of the distance-constraint reachability, considering the case where the threshold d is larger than the length of the 
longest path. In order to solve the DCR problem they  provide an estimator based on
a direct sampling approach and two new estimators based on unequal probability sampling and  recursive sampling~\cite{Jin11}.
Furthermore, they proposed  a divide and conquer exact algorithm that compute 
exact s-t DCR by  recursively partitioning all the possible discrete graphs from the probabilistic graph 
into groups so that the reachability of these groups can be computed easily. 

The need to model the uncertainty inherent in the data has increased  the attention on  \emph{probabilistic databases}.
In this framework exact approaches are infeasible for large database~\cite{Dalvi:2007:EQE:1285882.1285906} 
and hence the research has focused on computing approximate answers~\cite{Koch:2008:APE:1376916.1376932}.
An important probabilistic databases issue regards  the efficient evaluation of top-k queries.
A traditional  top-k query returns the $k$ objects with the maximum scores based on some scoring function.
In the uncertain world the scoring function becomes a probabilistic function. \cite{Soliman07top-kquery} formalized the  problem 
and \cite{Li:2011:UAR:1969331.1969353}  proposed a unified approach to ranking in probabilistic databases.

In this paper we adopt the probabilistic graphs framework to deal with uncertain problems exploiting 
both edges probabilistic values and edges labels 
denoting the type of relationships between two nodes.
Our work exploits the reachability tool using a direct sampling approach and considers as a constraint, 
instead of  the number of visited edges or the likelihood of the path,  the concatenation of the labels of the visited edges  going from a node to 
another.
We can consider the approach proposed in this paper as a generalization of the DCR problem
since we can consider homogeneous labels and a constraint length of the paths.

\section{CONCLUSIONS}

In this paper the \texttt{Eagle} system integrating  a framework based on probabilistic graphs able to deal with link prediction problems adopting
reachability  has been presented. We proposed a learning method to compute the most likely relationship between
two nodes in probabilistic graphs. In  particular,  we used a probabilistic graph in order to represent uncertain data and
relationships  and we adopted  the  reachability tool  to compute  the probability  of unknown 
  interconnections between two  nodes not directly connected. 
Each  of these connections may  be viewed  as probabilistic features and we can describe each observed link in the graph
as a feature vector. Given  the  training  set of observed links a L2-regularized Logistic Regression has  been  adopted to
 learn a  model  able to predict the label of  unobserved links. The application domain we chosen corresponds to the problem 
of recommender systems. The experimental evaluation proved that the proposed  approach achieves better results when compared to  that obtained  
with models induced by Singular Value Decomposition  on the user-item ratings matrix, representing one of the
best recent method for this kind of problem.

\bibliographystyle{amsplain}
\bibliography{ecai2012}
\end{document}